\documentclass[journal]{IEEEtran}
\usepackage{xcolor,soul,framed} 
\usepackage[colorinlistoftodos]{todonotes}
\colorlet{shadecolor}{yellow}
\usepackage[cmex10]{amsmath}
\usepackage{array}
\usepackage{mdwmath}
\usepackage{mdwtab}
\usepackage{eqparbox}
\usepackage{url}
\usepackage{tabularx}
\usepackage{amsfonts}
\usepackage{amssymb}
\usepackage{acronym}
\usepackage{verbatim}
\usepackage{xfrac,scalerel}
\newcommand\wh[1]{\hstretch{2}{\hat{\hstretch{.5}{#1}}}}
\usepackage{footnotebackref}

\usepackage{fixfoot}
\DeclareFixedFootnote{\repnote}{\url{http://computer.njnu.edu.cn/Lab/LABIC/LABIC_Software.html}}

\DeclareFixedFootnote{\secdnote}{\url{http://bailando.sims.berkeley.edu/enron_email.html}}

\begin{document}
\bstctlcite{IEEEexample:BSTcontrol}
    \title{Saliency-based Weighted Multi-label Linear Discriminant Analysis}
  \author{Lei~Xu\IEEEauthorrefmark{1},~\IEEEmembership{Student Member,~IEEE,}
      Jenni~Raitoharju\IEEEauthorrefmark{2},~\IEEEmembership{Member,~IEEE,}\\
      Alexandros~Iosifidis\IEEEauthorrefmark{3},~\IEEEmembership{Senior~Member,~IEEE,}
      and~Moncef Gabbouj\IEEEauthorrefmark{1},~\IEEEmembership{Fellow,~IEEE}
      \\
      \IEEEauthorblockA{\IEEEauthorrefmark{1}Department of Computing Sciences, Tampere University, Finland\\
\IEEEauthorrefmark{2}Programme for Environmental Information, Finnish Environment Institute, Finland\\
\IEEEauthorrefmark{3}Department of Engineering, Aarhus University, Denmark\\
Emails: \IEEEauthorrefmark{1}lei.xu@tuni.fi,
\IEEEauthorrefmark{2}jenni.raitoharju@environment.fi,
\IEEEauthorrefmark{3}ai@eng.au.dk,}
\IEEEauthorrefmark{1}moncef.gabbouj@tuni.fi}


\maketitle

\begin{abstract}
In this paper, we propose a new variant of Linear Discriminant Analysis (LDA) to solve multi-label classification tasks. The proposed method is based on a probabilistic model for defining the weights of individual samples in a weighted multi-label LDA approach. Linear Discriminant Analysis is a classical statistical machine learning method, which aims to find a linear data transformation increasing class discrimination in an optimal discriminant subspace. Traditional LDA sets assumptions related to Gaussian class distributions and single-label data annotations. To employ the LDA technique in multi-label classification problems, we exploit intuitions coming from a probabilistic interpretation of class saliency to redefine the between-class and within-class scatter matrices. The saliency-based weights obtained based on various kinds of affinity encoding prior information are used to reveal the probability of each instance to be salient for each of its classes in the multi-label problem at hand. The proposed Saliency-based weighted Multi-label LDA approach is shown to lead to performance improvements in various multi-label classification problems.
\end{abstract}


%
\IEEEpeerreviewmaketitle

\section{Introduction}
Multi-label classification tasks have become more and more common in the machine learning field recently, e.g., in text information categorization \cite{Li2015}, image and video annotation \cite{Qi2007}, sequential data prediction \cite{Read2017}, or music information retrieval \cite{Trohidis2011}. Multi-label databases exist for various real applications, such as Yeast database for protein localization sites prediction \cite{Nakai1992}, CAL500 database for music retrieval \cite{Turnbull2008}, or medical database for text classification \cite{Pestian2007}.

Compared to single-label problems, the characteristics of multi-label problems are more complicated and unpredictable. In a single label problem, each instance merely belongs to a specific class in a mutually exclusive manner \cite{Wang2010}. Classes in a multi-label problem are not mutually exclusive, which means that each data item can belong to either one or several classes. Moreover, different classes contain a varying number of data items, leading to class imbalanced problems \cite{Lu2019}. Hence, in order to solve a multi-label classification problem efficiently and effectively, we need not only to consider the correlation of class labels and features of each data item, but also take into account the different cardinalities of the classes. 

As described in \cite{Zhang2014}, multi-label classification methods are derived following either a problem transformation (PT) approach or an algorithm adaptation (AA) approach. Methods following the PT approach simply utilize single-label classification algorithms to tackle multi-label classification tasks using decomposition approaches, such as the binary relevance (BR) algorithm \cite{Tanaka2015}, \cite{ZhangM.L.LiY.K.LiuX.Y.Geng2018} or the label powerset (LP) algorithm \cite{Abdallah2016}, \cite{Tsoumakas2010}. A weighted multi-label linear discriminant analysis algorithm (wMLDA) \cite{Xu2018b} combines the decomposition approach with different labels and/or feature information to build a multi-label classification method. Methods following the AA approach directly utilize the information of class labels and data items to explore their correlation, e.g., in an extension of AdaBoost algorithm \cite{Schapire2000} or a de-convolution-based method in \cite{Streich2008}.  
Linear Discriminant Analysis (LDA) and its variants have been widely used to extract discriminant data representations for solving various problems involving supervised dimensionality reduction, e.g., in human action recognition \cite{Iosifidis2012b,Iosifidis2014b}, \cite{Wu2017}, biological data classification \cite{Wang2017,Huang2009a}, and facial image analysis \cite{Gao2009}. However, it cannot be directly used to tackle multi-label problems due to the characteristics of multi-label data. This is due to two factors: a) the contribution of each data item in the calculation of the scatter matrices involved in the optimization problem of single-label LDA and its variants cannot be appropriately determined and b) the cardinality of the various classes forming the multi-label problem can be quite imbalanced. 

In this paper, we propose a novel method for multi-label data classification based on a probabilistic approach that is able to estimate the contribution of each data item to the classes it belongs to by taking into account prior information encoded using various types of metrics. The proposed calculation of the contribution of each data item to the classes it belongs to cannot only weight its importance, but can also address problems related to imbalanced classes. To this end, we exploit the concept of class saliency introduced in \cite{Xu2018c}. Hence, the proposed method is named as Saliency-based Weighted Multi-label Linear Discriminant Analysis (SwMLDA). Our proposed SwMLDA approach, as a kind of PT approach, exploits both label and feature information with various prior weighting factors, i.e., binary-based weight form \cite{Park2008a}, misclassification-based weight form \cite{Xu2018c}, entropy-based weight form \cite{Chen2007}, fuzzy-based weight form \cite{Lin2010}, dependence-based weight form \cite{Xu2018b}, and correlation-based weight form \cite{Wang2010}. The proposed method leads to improved results on 10 publicly available multi-label databases. 

We have made the following contributions on multi-label classification tasks with our novel SwMLDA approach: (1) we propose using probabilistic saliency estimation in multi-label classification to weight the importance of each item for its classes; (2) we formulate a novel SwMLDA method that uses the saliency-based weights and can alleviate the problems related to imbalanced datasets; (3) we integrate label and feature information to SwMLDA by using various types of weighting factors as prior information; (4) we compare our proposed approach to related methods on 10 diverse multi-label data sets, and the results show considerable improvements in multi-label classification tasks using our approach. 

The remainder of this paper is structured as follows. In Section 2, we briefly review the related works. We include a precise explanation of the LDA and weighted MLDA with adequate mathematical notations to support the derivations of the probabilistic saliency estimation. In Section 3, we describe our proposed methods in detail. Section 4 presents for experimental setup and results on 10 multi-label databases. In Section 5, we conclude this paper and discuss the potential future studies.

\section{Related works}
In this section, we first briefly present several standard approaches for multi-label classification in subsection \ref{SS:PriorMLC}. In subsection \ref{SS:DimRedMLC}, we provide a detailed description of the standard LDA, weighted LDA, Multi-label LDA (MLDA), and weighted Multi-label LDA (wMLDA), since they form the theoretical foundation for the proposed work. Subsequently, we introduce the general concepts of saliency estimation and the probabilistic saliency estimation approach needed to develop the proposed method.

\subsection{General methods for multi-label classification tasks}\label{SS:PriorMLC}
Various methods have been proposed for solving multi-label classification tasks, such as variants of Support Vector Machine (SVM) \cite{Godbole2004} and various feature extraction methods \cite{Wang2010,Xu2018b,Zhang2008}. As PT algorithms, Binary Relevance-based methods \cite{Tanaka2015,ZhangM.L.LiY.K.LiuX.Y.Geng2018,Read2009} decompose a multi-label classification problem into several single-label classification problems in a one-versus-all manner. Another standard PT method is Label Powerset (LP) algorithm \cite{Abdallah2016}, \cite{Pushpa2017}, which exploits the dependencies or correlations of class labels to rebuild a labeled subset for a single-label classifier. The traditional SVM algorithm acts as a PT approach: in \cite{Boutell2004} a multi-label scene classification problem is decomposed into several single-label problems by following a cross-training strategy. 

As an AA approach, alternating decision-tree (ADTree) was proposed to enhance the performance of boosting methods \cite{Yoav1999,DeComite2003}. In \cite{Yoav1999}, the strategy of an alternating decision tree is based on an option tree using boosting. Another decision-tree related algorithm ADTboost.MH was proposed in \cite{DeComite2003} to solve multi-label text and data classification problems by ADTboost algorithm\cite{Yoav1999} and Adaboost.MH algorithm \cite{Schapire1999}. 

\subsection{Dimensionality reduction algorithms for multi-label classification tasks}\label{SS:DimRedMLC}
Standard LDA and its variants have been applied to tackle various multi-label classification problems \cite{Wang2010, Park2008a,Oikonomou2013,Yuan2014,Nie2009,Siblini2019}. Generally, dimensionality reduction-based methods tackling multi-label classification problems are categorized as unsupervised and supervised, depending on whether class label information is involved in or not \cite{Xu2018b}. The objective of dimensionality reduction-based methods is to determine a data projection matrix $\mathbf{W}\in\mathbb{R}^{D \times d}$ mapping the data from the original feature space $\mathbb{R}^D$ to a discriminant subspace $\mathbb{R}^d$, where $D > d$. 

\subsubsection{Linear Discrimination Analysis}
LDA is an effective technique to reduce dimensionality of original data as a prepossessing step for single-label classification problems. In the following, we assume that a training set formed by $N$ data points and class labels is presented as
\begin{equation} \label{eq: data_sets}
    \{(\mathbf{x}_1, \mathbf{y}_1), ..., (\mathbf{x}_i, \mathbf{y}_i), ..., (\mathbf{x}_N, \mathbf{y}_N)\},
\end{equation}
where $\mathbf{x}_i \in \mathbb{R}^D$ and $\mathbf{y}_i \in \mathbb{R}^C$ are the data points and the corresponding label vectors, respectively. The instance matrix $\mathbf{X} \in \mathbb{R}^{D \times N}$ is defined as 
\begin{equation} \label{eq: multi_label_vector}
    \mathbf{X} = [\mathbf{x}_1, ..., \mathbf{x}_i, ..., \mathbf{x}_N].
\end{equation}

The label matrix $\mathbf{Y} \in \mathbb{R}^{C \times N}$ is depicted as
\begin{equation} \label{eq: single_label_vector}
    \mathbf{Y} = [\mathbf{y}_{1}, ..., \mathbf{y}_{i}, ..., \mathbf{y}_{N}] = [\mathbf{y}_{(1)}, ..., \mathbf{y}_{(j)}, ..., \mathbf{y}_{(C)}]^\intercal.
\end{equation}
The label information of element $\mathbf{x}_i = [x_{1i}, ..., x_{ji}, ..., x_{Di}]^\intercal$ is represented as $\mathbf{y}_{i} = [y_{1i}, ..., y_{ji}, ..., y_{Ci}]^\intercal$. If $\mathbf{x}_i$ belongs to the class $c \in \{1, ..., C\}$, $y_{ci} = 1$, otherwise $y_{ci} = 0$. Note that in single label-classification tasks there is a single 1 on each column. Later, we will use the same notation in multi-label classification, where the number of 1s is not constrained.

The within-class, between-class, and total scatter matrices $\mathbf{S}_w$, $\mathbf{S}_b$, and $\mathbf{S}_t$, respectively, are defined as follows:
\begin{equation} \label{eq:S_W}
    \mathbf{S}_w = \sum_{c=1}^{C} \sum_{i=1}^{N} y_{ci} (\mathbf{x}_i - \mbox{\boldmath$\mu$}_c)(\mathbf{x}_i - \mbox{\boldmath$\mu$}_c)^T,
\end{equation}
\begin{equation} \label{eq:S_B}
    \mathbf{S}_b = \sum_{c=1}^{C} (\sum_{i=1}^{N} y_{ci}) (\mbox{\boldmath$\mu$}_c - \mbox{\boldmath$\mu$})(\mbox{\boldmath$\mu$}_c - \mbox{\boldmath$\mu$})^T,
\end{equation}
\begin{equation} \label{eq:S_B}
    \mathbf{S}_t = \sum_{c=1}^{C} \sum_{i=1}^{N} y_{ci} (\mathbf{x}_i - \mbox{\boldmath$\mu$})(\mathbf{x}_i - \mbox{\boldmath$\mu$})^T.
\end{equation}
$\mbox{\boldmath$\mu$}_c$ denotes the mean vector of class $c$ as
\begin{equation} \label{eq:mu_c}
    \mbox{\boldmath$\mu$}_c = \frac{1}{N_c} \sum_{i=1}^{N} y_{ci} \mathbf{x}_i,
\end{equation}
where $N_c = \sum_{i=1}^{N} y_{ci}$ is the cardinality of class $c$.
The total mean vector $\mbox{\boldmath$\mu$}$ is computed as
\begin{equation} \label{eq: mu}
    \mbox{\boldmath$\mu$} = \frac{1}{N} \sum_{i=1}^N \mathbf{x}_i.
\end{equation}

The optimal projection matrix $\mathbf{W}$ is learned by maximizing the Fisher's discriminant criterion \cite{R.A.Fisher1936} through compacting the within-class scatter and maximizing the between-class scatter simultaneously as
\begin{equation} \label{eq:Fisher's optimization}
    J(\mathbf{W}) = \underset{\mathbf{W}}{\text{argmax}} \:\: \frac{\text{tr}(\mathbf{W}^T \mathbf{S}_b \mathbf{W})}{\text{tr}(\mathbf{W}^T \mathbf{S}_w \mathbf{W})},
\end{equation}
where $\text{tr}(.)$ denotes the trace of a matrix. Usually, the optimal projection matrix $\mathbf{W}$ is calculated by solving eigenvalue decomposition of the matrix $\mathbf{S} = \mathbf{S}_{w}^{-1}\mathbf{S}_{b}$ and then using the eigenvectors corresponding to the largest eigenvalues as the projection matrix $\mathbf{W}$. The rank of $\mathbf{S}$ is equal to $C - 1$, which is the maximal dimensionality of the resulting subspace. Since $\mathbf{S}_t = \mathbf{S}_w + \mathbf{S}_b$, an alternative approach is to use $\mathbf{S}_t$ instead of $\mathbf{S}_w$ and maximize the Fisher's discriminant criterion as
\begin{equation} \label{eq:fisher_wt}
J(\mathbf{W}) = \underset{\mathbf{W}}{\text{argmax}} \frac{\text{tr}(\mathbf{W}^T \mathbf{S}_b \mathbf{W})}{\text{tr}(\mathbf{W}^T \mathbf{S}_t \mathbf{W})}.
\end{equation}

Although the traditional LDA technique has gained popularity on various single-label classification tasks, its performance varies according to the types of input data sets. Usually, the data sets used in most traditional LDA classification tasks are assumed to have equal class distribution as a homoscedastic Gaussian model \cite{Petridis2004a}, in which the covariance matrices of each class should be identical \cite{Tang2005}. Furthermore, the performance is affected severely due to the imbalance of input data sets \cite{Tang2005a}.

\subsubsection{Weighted Linear Discrimination Analysis}
In order to enhance the robustness of traditional LDA on different kinds of data sets, various weight factors are introduced into the definitions of scatter matrices to balance the contribution of each class, according to class statistics \cite{Tang2005}, \cite{Li2009a}, e.g., class cardinality, a prior probability. Weighted LDA approaches have diminished the influence of outlier classes on the scatter matrices of imbalanced data sets to some extent; however, they still neglect the varying importance of individual samples in the class description. Saliency-based weighted Linear Discriminant Analysis (SwLDA) \cite{Xu2018c}, as a kind of graph expression, was proposed to explore the contribution of each instance based on probabilistic saliency estimation \cite{Aytekin2018}. Our work uses the same idea for multi-label classification.

Generally, weight factors are calculated using various metrics to reallocate the contribution of each class, which can alleviate the influence of outlier classes on the projection matrix. An example of a weighted between-class matrix definition based on Bayes error rate was proposed in \cite{Loog2001b}: 
\begin{equation} \label{eq:4}
\mathbf{S}_b = \sum_{k=1}^{C-1}\sum_{l=k+1}^{C}L_{kl} p_k p_l (\mbox{\boldmath$\mu$}_k - \mbox{\boldmath$\mu$}_l)(\mbox{\boldmath$\mu$}_k - \mbox{\boldmath$\mu$}_l)^T,
\end{equation}
 where $p_k$, $p_l$ denote the a priori probabilities of class $k$ and class $l$, respectively. $L_{kl}$ expresses the dissimilarity between class $k$ and class $l$. The within-class scatter matrix can be muted with prior information as in \cite{Tang2005}:
 \begin{equation} \label{eq:5}
\mathbf{S}_w = \sum_{c=1}^{C} \sum_{i=1}^{N} y_{ci} r_c (\mathbf{x}_i - \mbox{\boldmath$\mu$}_c) (\mathbf{x}_i - \mbox{\boldmath$\mu$}_c)^T,
\end{equation}
where $r_c$ is a relevance weight factor that has a low value if class $c$ is estimated to be an outlier class. Thus, both definitions of scatter matrices decrease the influence of outlier classes. After computing the weighted scatter matrices, they can be used to obtain the optimal projection matrix $\mathbf{W}$ from Eq. (\ref{eq:Fisher's optimization})

\subsubsection{Multi-label Linear Discrimination Analysis}
\begin{figure}
  \includegraphics[width=8.5cm,height=5.5cm]{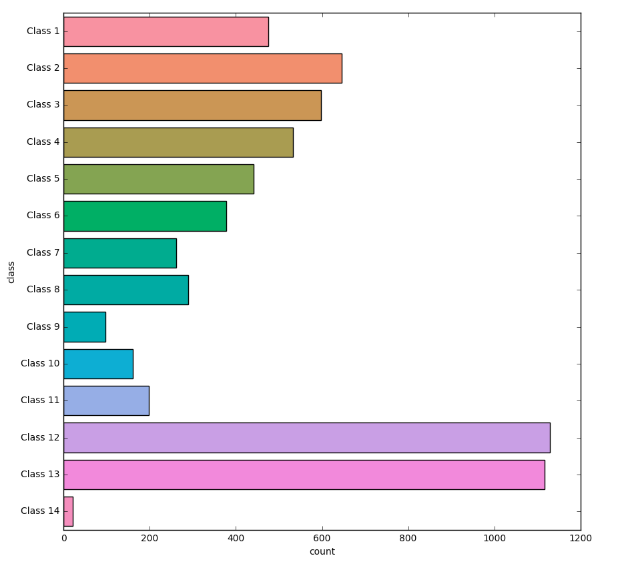}
  \caption{The number of instances for each class in Yeast database}
  \label{yeast_data}
\end{figure}

Although weighted LDA algorithms have enhanced the performance on single-label classification tasks \cite{Jarchi2006a}, \cite{Ahmed2012} compared to traditional LDA, such variants are still not directly applicable for multi-label classification tasks \cite{Wang2010}. In a multi-label data set, label information contains certain correlations or dependencies \cite{Wu2016}, for example, an image instance labeled as 'car' highly correlates to label 'road' \cite{Wang2010}. Besides, it is quite common that the number of samples in each class in a multi-class data set is imbalanced. For example, the largest class size is 1128 and the smallest 21 in the widely used Yeast database \cite{Nakai1992}, as shown in Fig. \ref{yeast_data}. Due to the specific characteristics of multi-label databases, it is imperative to take into account the correlation of class labels and/or discriminative feature information of each instance to tackle the sub-optimal classification result on imbalanced data sets. 

When traditional LDA and its variants are applied to tackling multi-label classification tasks by simply using Eqs. (\ref{eq:S_W}) - (\ref{eq:S_B}) with the multi-label label matrix $\mathbf{Y}$, a significant problem is that the contribution of one instance can be repeatedly counted in computing the scatter matrices. Hence, weight factors are used to express redundancy or/and correlation information so that LDA related algorithms can calculate scatter matrices without redundancy on multi-label databases. In \cite{Wang2010}, a multi-label linear discriminant analysis (MLDA) approach based on the exploration of label correlation information was proposed to tackle multi-label image or video classification tasks. MLDA approach embeds the correlation information of class labels as weight factors in the definition of scatter matrices as
\begin{equation} \label{eq:S_w}
\mathbf{S}_w = \sum_{c=1}^{C} \sum_{i=1}^{N} p_{ci} (\mathbf{x}_i - \mbox{\boldmath$\mu$}_c) (\mathbf{x}_i - \mbox{\boldmath$\mu$}_c)^T,
\end{equation}

\begin{equation} \label{eq:S_b}
\mathbf{S}_b = \sum_{c=1}^{C} \Big( \sum_{i=1}^{N} p_{ci} \Big) (\mbox{\boldmath$\mu$} - \mbox{\boldmath$\mu$}_c) (\mbox{\boldmath$\mu$} - \mbox{\boldmath$\mu$}_c)^T,
\end{equation}

\begin{equation} \label{eq:S_t}
\mathbf{S}_t = \sum_{c=1}^{C} \sum_{i=1}^{N} p_{ci} (\mathbf{x}_i - \mbox{\boldmath$\mu$}) (\mathbf{x}_i - \mbox{\boldmath$\mu$})^T,
\end{equation}
where $p_{ci}$ describes a weight factor of the $i^{th}$ instance for the class $c$, $ \mbox{\boldmath$\mu$}$ is the total mean vector of all training instances, and $\mbox{\boldmath$\mu$}_c$ is the mean vector of class $c$:
\begin{equation} \label{eq:mu_m}
    \mbox{\boldmath$\mu$} = \frac{\sum_{c=1}^{C}\sum_{i=1}^{N}p_{ci}\mathbf{x}_{i}}{\sum_{c=1}^{C}\sum_{i=1}^{N}p_{ci}},
\end{equation}
\begin{equation} \label{eq:mu_c}
    \mbox{\boldmath$\mu$}_c = \frac{\sum_{i=1}^{N}p_{ci}\mathbf{x}_{i}}{\sum_{i=1}^{N}p_{ci}}.
\end{equation}

A correlation matrix $\mathbf{R} \in \mathbb{R}^{C \times C}$ is computed using the class labels of each pair of classes:
\begin{equation} \label{eq:C_t}
R_{kl} = \text{cos}(\mathbf{y}_{(k)}, \mathbf{y}_{(l)}) = \frac{\langle \mathbf{y}_{(k)}, \mathbf{y}_{(l)} \rangle }{\| \mathbf{y}_{(k)}\| \| \mathbf{y}_{(l)}\|},
\end{equation}
where $\mathbf{y}_{(k)}$, $\mathbf{y}_{(l)}$ are label vectors for classes $k, l \in {1, ..., C}$. The label correlation information can reveal whether two classes are closely related or not. The correlation matrix $\mathbf{R}$ is then used to compute the weight factors in Eqs. (\ref{eq:S_w}) - (\ref{eq:mu_c}). To tackle the over-counting problem \cite{Wang2010}, the weight factors are normalized with $\ell_1$-norm:
\begin{equation} \label{eq:corr_P}
\mathbf{v}_i  = \frac{\mathbf{R} \; \mathbf{y}_i}{\| \mathbf{y}_i\|_{\ell_1}},
\end{equation}
where $\mathbf{y}_i,\; i \in \{1, ..., N\}$ is the label vector for the $i^{th}$ sample. $\mathbf{v}_i$ was used directly as $\mathbf{p}_i$ in \cite{Wang2010}, while we exploit it in a different manner in our work.

Various weight matrices have been introduced to improve the performance of LDA on multi-label classification tasks \cite{Wang2010}, \cite{Xu2018b}, \cite{Oikonomou2013}. Such strategies yield a more suitable projection subspace compared to other dimensionality reduction algorithms \cite{Wang2010}, such as principle component analysis (PCA), multi-label dimensionality reduction via dependence maximization (MDDM), or multi-label least square (MLLS). 

In \cite{Oikonomou2013}, MLDA was extended to Direct MLDA by changing the definition of $\mathbf{S}_b$ in a way that allows to obtain a higher dimensional subspace than the original MLDA, where the subspace dimensionality is limited by the rank of $\mathbf{S}_b$ to $C-1$. This extension work further enhanced the results in multi-label video classification tasks. Another extension, multi-label discriminant analysis with locality consistency (MLDA-LC) \cite{Yuan2014} not only preserves the global class label information as MLDA does, but also incorporates a graph regularized term to utilize the local geometric information. MLDA-LC reveals the similarity among nearby instances with transformation in the projection space using incorporation of the graph Laplacian matrix into the MLDA approach, which further enhances the classification performance in multi-label data sets compared to MLDA or/and MLLS algorithms. 

\subsubsection{Weighted multi-label linear discriminant analysis}
A weighted multi-label LDA (wMLDA) approach was proposed in \cite{Xu2018b} focusing on linear feature extraction for multi-label classification. In wMLDA, a multi-label classifier is composed of several single-label classifiers according to the number of classes and a weight matrix is simultaneously calculated based on various metrics to embody the contribution of each instance in scatter matrices calculation. Various metrics can be used to measure the relationships among instances from the labels and/or features. wMLDA approach employs correlation-based weight form \cite{Wang2010}, entropy-based weight form \cite{Chen2007}, binary-based weight form \cite{Park2008a}, fuzzy-based weight form \cite{Lin2010}, and dependence-based weight form \cite{Xu2018b}. In this work, we exploit the same metrics, while we use this information in a novel way. We provide a detailed explanation of the metrics in Section \ref{prior_info}.

In \cite{Xu2018b}, scatter matrices $\mathbf{S}_w$, $\mathbf{S}_b$, and $\mathbf{S}_t$ are redefined to exploit the prior information for weighting. Firstly, a non-negative weight matrix $\mathbf{P} \in \mathbb{R}^{C \times N}$ with the same size of label matrix $\mathbf{Y}$ is defined to describe the weight of each instance to its corresponding classes:
\begin{equation} \label{eq:weights}
    \mathbf{P} = [\mathbf{p}_1, ..., \mathbf{p}_i, ..., \mathbf{p}_N] = [\mathbf{p}_{(1)}, ..., \mathbf{p}_{(j)}, ..., \mathbf{p}_{(C)}]^\intercal ,
\end{equation}
where $\mathbf{p}_i$ represents a weight vector for the $i^{th}$ instance and $\mathbf{p}_{(j)}$ is a weight vector for the $j^{th}$ class. The weight matrix $\mathbf{P}$ is calculated based on one of the prior information matrices described in \ref{prior_info}. Then, $n_c$ and $n$ are defined as summations of weights for the $c^{th}$ class and all classes:
\begin{equation} \label{eq:sum_nc}
    n_c = \sum_{i=1}^{N} p_{ci}, \:\:\:\:\: c = 1, ..., C, 
\end{equation}
\begin{equation} \label{eq:sum_nc}
    n = \sum_{c=1}^{C} \sum_{i=1}^{N}  p_{ci} = \sum_{c=1}^{C} n_c.
\end{equation}
In order to simplify notation, row vectors $\hat{\mathbf{n}}$ and $\hat{\mathbf{p}}$ are defined as
\begin{equation} \label{eq:n_vec}
    \hat{\mathbf{n}} = \bigg[ \frac{1}{n_1}, ..., \frac{1}{n_C} \bigg].
\end{equation}
\begin{equation} \label{eq:sum_weig}
    \hat{\mathbf{p}} = [ \hat{p}_1, ..., \hat{p}_i, ..., \hat{p}_N] = \sum_{c=1}^{C} \mathbf{p}_{(c)},
\end{equation}
where $\hat{p}_i$ is the summation of weights for the $i$th instance over all classes $\hat{p}_i = \sum_{c=1}^{C}p_{ci}, i = 1, ..., N$. Then, the scatter matrices can be redefined as
\begin{equation} 
    \begin{split}  \label{eq:S_w_m}
    \mathbf{S}_w & =  \sum_{c=1}^{C} \sum_{i=1}^{N} p_{ci} (\mathbf{x}_i - \mbox{\boldmath$\mu$}_c) (\mathbf{x}_i - \mbox{\boldmath$\mu$}_c)^T  \\
                 & = \mathbf{X} \Big(\text{diag}(\wh{\mathbf{p}}) - \wh{\mathbf{P}}^\intercal \wh{\mathbf{P}}\Big) \mathbf{X}^\intercal,
    \end{split}
\end{equation}

\begin{equation} 
    \begin{split}\label{eq:S_b_m}
    \mathbf{S}_b & = \sum_{c=1}^{C} \Big( \sum_{i=1}^{N} p_{ci} \Big) (\mbox{\boldmath$\mu$} - \mbox{\boldmath$\mu$}_c) (\mbox{\boldmath$\mu$} - \mbox{\boldmath$\mu$}_c)^T \\
                & = \mathbf{X} \Big(\wh{\mathbf{P}}^\intercal \wh{\mathbf{P}} - \frac{1}{n}\wh{\mathbf{p}}^\intercal \wh{\mathbf{p}} \Big) \mathbf{X}^\intercal.
    \end{split}
\end{equation}

\begin{equation} \label{eq:S_t_m}
    \begin{split}
        \mathbf{S}_t & = \mathbf{S}_w + \mathbf{S}_b \\
                     & = \mathbf{X} \Big(\text{diag}(\wh{\mathbf{p}}) - \frac{1}{n}\wh{\mathbf{p}}^\intercal \wh{\mathbf{p}} \Big) \mathbf{X}^\intercal,
    \end{split}
\end{equation}
where $\wh{\mathbf{P}} = \mathbf{P} \text{diag} (\wh{\mathbf{n}}^{\frac{1}{2}})$ has row vectors $\frac{\mathbf{p}_{(c)}}{\sqrt{n_{c}}} \: (c = 1, ..., C)$. Under this approach, the optimal projection matrix $\mathbf{W}$ can still be obtained by solving the generalized eigenproblem corresponding to Eq. (\ref{eq:fisher_wt}) as discussed in Section 2. 

\subsection{Saliency estimation} \label{sec:2C}
Saliency estimation as a standard computer vision task is inspired by neurobiological studies \cite{Ltti1998} and cognition psychology \cite{Treisman1980}. Generally saliency estimation is a pre-processing step for various high-level computer vision tasks, such as object detection \cite{Aytekin2018}, \cite{WangSalientSurvey}, omni-directional images \cite{Battisti2018}, and human attention estimation \cite{Choi2016HumanApproach}. Saliency in physiological science is defined as a special kind of perception of the human visual system, by which humans can perceive particular parts in a scene in details due to colors, textures, or other prominent information contained in these parts \cite{Cheng2011}. These particular parts can be distinguished as foreground from non-salient background parts. 

Computational saliency estimation approaches can be categorized as local approaches and global approaches based on the way they process saliency information \cite{Cheng2011}. Local saliency approaches explore the prominent information around the neighborhood of specific pixels/regions whilst global approaches exploit the rarity of a pixel/patch/region in the whole scene. Since the emergence of computational saliency estimation field in \cite{Koch1985}, various probabilistic approaches have been explored in this topic. In \cite{Jian2018}, a saliency map is estimated based on three kinds of prior information on images at super-pixel level. Saumya et al. utilize a generalized Bernoulli distribution to estimate a saliency map in their work \cite{Jetley2016}. 

Another saliency estimation approach was proposed by Aytekin et al. \cite{Aytekin2018} for segmenting salient objects in an image using a probabilistic estimation, where a probability mass function $\mathbf{P}(\mathbf{x})$ depicts whether a region $\mathbf{x}_i$ (pixel, super-pixel, or patch) in an image is considered as a distinct region. The higher the values of $\mathbf{P}(\mathbf{x})$ for a region, the more prominent the region is. $\mathbf{P}(\mathbf{x})$ is solved by optimizing two items working simultaneously to allocate not only lower probability to non-salient regions but also similar probabilities to similar regions as follows: 
\begin{equation} 
    \begin{split}  \label{eq:pse}
        &\underset{\mathbf{P}(x)}{\text{argmin}}\bigg(\sum_i(\mathbf{P}(\text{x} = \mathbf{x}_i)\big)^2 v_i  \\
        &+ \bigg(\sum_{i,j}\Big(\big(\mathbf{P}(\text{x} = \mathbf{x}_i)\big)^2-\mathbf{P}(\text{x} = \mathbf{x}_i)\mathbf{P}(\text{x} = \mathbf{x}_j) \Big)w_{ij}\bigg) \bigg) \\
        & \text{s.t.} \quad \sum_i\mathbf{P}(\text{x} = \mathbf{x}_i) = 1,  
    \end{split}
\end{equation}
where the first term suppresses the probability of a non-prominent region $i$ using its prior information $v_i \ge 0$. In the second term, a high similarity of regions $i$ and $j$, given as a high similarity value $w_{ij}$, forces the regions to have similar probabilities. 

This optimization task in Eq. (\ref{eq:pse}) can be expressed using matrix notations as
\begin{gather}
    \mathbf{p^*} = \underset{\mathbf{p}}{\text{argmin}} \:\: (\mathbf{p^T H p}), \label{eq:14}\\
    \mathbf{H} = \mathbf{D} - \mathbf{W} + \mathbf{V}, \label{eq:15} \\
\text{s.t.}\quad \mathbf{p^T1} = 1,\nonumber
\end{gather}
\noindent
where $\mathbf{p}$ is a probability vector that depicts the probabilities of each element or region $i$ to be salient, i.e., $p_i = \mathbf{P}( \text{x} = x_i)$. $\mathbf{W}$ is an affinity matrix, which denotes the similarity of each pair of regions $i$ and $j$ as $[\mathbf{W}]_{ij} = w_{ij}$. $\mathbf{D}$ is a diagonal matrix having elements equal to $[\mathbf{D}_{ii}] = \sum_j w_{ij}$. $\mathbf{V}$ is a diagonal matrix having elements $[\mathbf{V}]_{ii} = v_i$. Then Lagrangian multiplier method is employed
\begin{equation} \label{eq:pse_tmp}
\mathcal{L}(\mathbf{p}, \gamma) = (\mathbf{p^T H p}) - \gamma (\mathbf{p^T1} - 1).
\end{equation}
A global optimum $\mathbf{p^*}$ is obtained by setting the partial derivative of Eq. (\ref{eq:pse_tmp}) with the respect $\mathbf{p}$ to zero. The final optimized probability vector is
\begin{equation} \label{eq:final_pse}
\mathbf{p}_{pse}^* = \mathbf{H}^{-1} \mathbf{1}.
\end{equation}

\section{PROPOSED METHOD}
We propose a novel saliency-based weighted linear discriminant analysis method for multi-label classification tasks, where the saliency-based weight factors are calculated based on the probabilistic saliency estimation approach and the specific prior information of the input data. In this section, we describe our novel Saliency-based weighted Multi-label Linear Discriminant Analysis (SwMLDA) approach in detail. 

We calculate a saliency-based weight matrix $\mathbf{P} \in \mathbb{R}^{C \times N}$ based on the probabilistic saliency estimation with the exploration of various prior information: binary \cite{Park2008a}, correlation \cite{Wang2010}, entropy \cite{Chen2007}, fuzzy \cite{Lin2010}, dependence \cite{Xu2018b}, and misclassification \cite{Xu2018c}. The weight matrix $\mathbf{P}$ is denoted as:
\begin{equation} \label{eq:weights_*}
    \mathbf{P} = [\mathbf{p}_1, ..., \mathbf{p}_i, ..., \mathbf{p}_N] = [\mathbf{p}_{(1)}, ..., \mathbf{p}_{(j)}, ..., \mathbf{p}_{(C)}]^\intercal,
\end{equation}
where $\mathbf{p}_i \in \mathbb{R}^C$ represents the optimal weight vector of the $i^{th}$ instance and $\mathbf{p}_{(j)} \in \mathbb{R}^N$ is the weight vector of the $j^{th}$ class. The details for computing the probabilistic weight matrix $\mathbf{P}$ are given in the next subsections. After forming $\mathbf{P}$, the proposed method proceeds as wMLDA by using the weights in the scatter matrices $\mathbf{S}_b$ and $\mathbf{S}_t$:
\begin{equation} 
\label{eq:S_b_m}
    \mathbf{S}_b  = \mathbf{X} \Big({{\mathbf{P}}}^\intercal \; {\mathbf{P}} - \frac{1}{n}{\wh{\mathbf{p}}}^\intercal {\wh{\mathbf{p}}}\Big) \mathbf{X}^\intercal,
\end{equation}

\begin{equation} \label{eq:S_t_m}
\mathbf{S}_t = \mathbf{X} \Big(\text{diag}(\wh{\mathbf{p}}) - \frac{1}{n}{\wh{\mathbf{p}}}^\intercal\wh{\mathbf{p}}\Big) \mathbf{X}^\intercal,
\end{equation}
where $\wh{\mathbf{p}} = \sum_{c=1}^{C} \mathbf{p}_{(c)}$. Note that our method normalizes the weight vectors so that the sum of the weights for  a class is always 1. Therefore, $\hat{\mathbf{n}}$ in Eq. (\ref{eq:n_vec}) is an identity vector and $\hat{\mathbf{P}} = \mathbf{P}$. Finally, the optimal projection matrix $\mathbf{W}$ is obtained from Eq. (\ref{eq:fisher_wt}) by solving the corresponding generalized eigenvalue problem.

\subsection{Saliency-based weight factors}
We extend the probabilistic saliency estimation approach \cite{Aytekin2018} described in Section \ref{sec:2C} to express the saliency of each instance for its class(es). To this end, we formulate the prior information in $\mathbf{V}$ and $\mathbf{W}$ so that probability $\mathbf{p}_{(c)}$ describes the saliency of each item for class $c$. In an initial work \cite{Xu2018c}, we used saliency-based weight factors to tackle sub-optimal classification results caused by imbalanced data sets or/and outliers in single-label classification using LDA-based algorithms. Here, we exploit the saliency-based weight factors to tackle multi-label classification tasks. 

We calculate the saliency-based weight factors $\mathbf{p}^*_{(c)}$ separately for each class in the spirit of PT approaches. For each class, we consider only the samples belonging to the class, thus, $\mathbf{p}^*_{(c)}$ has $N_c$ elements. $\mathbf{p}^*_{(c)}$ is computed following Eq. (\ref{eq:final_pse}) as

\begin{equation} \label{eq:p_c}
\mathbf{p}^*_{(c)} = \mathbf{H}_c^{-1} \mathbf{1}, 
\end{equation}
where $\mathbf{H}_c$ constitutes three terms as $\mathbf{H}_c = \mathbf{D}_c - \mathbf{W}_c + \mathbf{V}_c$. To form $\mathbf{p}_{(c)}$ from $\mathbf{p}^*_{(c)}$, the elements of $\mathbf{p}^*_{(c)}$ are placed on their corresponding positions in $\mathbf{p}_{(c)}$ and the values for items not belonging to class $c$ are set to zero. We then form the weight matrix $\mathbf{P} \in \mathbb{R}^{C \times N}$ by placing weight vectors $\mathbf{p}_c$ as its rows.

$\mathbf{W}_c$ is an affinity matrix obtained by a graph notation. That is, for each class $c$, we form its corresponding graph $\mathcal{G}_C = \{\mathbf{X}_c, \mathbf{W}_c\}$, where $\mathbf{X}_c \in \mathbb{R}^{D \times N_c}$ is a matrix formed by the instances belonging to class $c$ and $\mathbf{W}_c \in \mathbb{R}^{N_c \times N_c}$ is a graph weighting matrix expressing the similarity between each pair of instances in class $c$. In our experiments, we use a fully connected graph to obtain $\mathbf{W}_c$ with a heat kernel function formulated as
\begin{equation} \label{eq:W_ij}
[\mathbf{W}_{c}]_{ij} = \exp\left(-\frac{\|\mathbf{x}_i - \mathbf{x}_j\|}{ 2 \sigma^2}\right),
\end{equation}
where $\mathbf{x}_i$ and $\mathbf{x}_j$ are the $i^{th}$ and $j^{th}$ instance in class $c$, and $i,j \in \{1, 2, ..., N_c\}$. $\sigma$ is set as a constant value. $\mathbf{D}_c$ is a diagonal matrix and each element is calculated based on $\mathbf{W}_c$ as $[\mathbf{D}_{c}]_{ii} = \sum_j [\mathbf{W}_{c}]_{ij}$.

$\mathbf{V}_c \in \mathbb{R}^{N_c \times N_c}$ is a diagonal matrix, which carries the prior information of each instance in class $c$ to be salient for its class based on the metrics presented in the next subsection. The values of $\mathbf{V}_c$ inversely relate to the values of weight factor vector $\mathbf{p^*}_{(c)}$ ranging from 0 to 1. The lower a value $[\mathbf{V}_c]_{ii}$, the more prominent the corresponding instance is expected to be based on the prior knowledge. We introduce six different prior information matrices to exploit label or/and feature information of each class, which produce six different variants of the proposed approach.

After computing the prior information vector $\mathbf{V}_c$ and affinity matrix $\mathbf{W}_c$ for class $c$, we follow the approach of PSE in Eq. (\ref{eq:p_c}) to calculate the saliency score vector ${\mathbf{p}^*}_{(c)}$ for class $c$. In order to avoid singularity during this process, a regularized version of $\mathbf{H}_c$ with a small value epsilon added to the diagonal elements is used. The summation of the values in the saliency-based weight vector ${\mathbf{p}^*}_{(c)}$ for each class is one, which is expected to alleviate the over-counting problem.

\subsection{Prior information matrices}\label{prior_info}

    \subsubsection{\textit{Misclassification-based prior information matrix (SwMLDAm)}} This approach was defined in \cite{Xu2018c} to alleviate the sub-optimal result in LDA arising from outlier instances on imbalanced data sets. We utilize the misclassification-based prior information to generate a diagonal matrix $\mathbf{V}_c$ based on the probability of each instance $i$ belonging to class $c$ to be more salient for another class:
    \begin{eqnarray}\label{eq:mis_classification}
          \begin{array}{r@{}l}
            [\mathbf{V}_c]_{ii} =
          \end{array}
            \left\lbrace
          \begin{array}{ll}
            0,  &\text{if} \:\:d_{ic}^c < \underset{k\neq c}{\text{min}} \:\:d_{ic}^k , \\
            \frac{ d_{ic}^c }{ \underset{k\neq c}{\text{min}} \:\:d_{ic}^k },      & \text{otherwise},
          \end{array}
          \right.
        \end{eqnarray}
        where $d_{ic}^k = \|\mathbf{x}_{ic} - \mbox{\boldmath$\mu$}_k\|^2_2$, $\mathbf{x}_{ic}$ is the $i^{th}$ instance of class $c$ and $\mbox{\boldmath$\mu$}_k$ is the mean vector of class $k$. In this approach, a sample which is closer to another class is considered less salient for class $c$ even if it is relatively close to the center of class $c$. 
        
     \subsubsection{\textit{Correlation-based prior information matrix (SwMLDAc)}} As in \cite{Wang2010}, label correlation information is represented by a class pair matrix $\mathbf{R}$ defined in Eq. (\ref{eq:C_t}). For each instance $i$, the normalized weight vector $\mathbf{v}_i \in \mathbb{R}^C$ is calculated by Eq. (\ref{eq:corr_P}). We compute the weight factors separately for each class $c$ and, after obtaining them for all instances, we select the $c^{th}$ elements, $v_{ic}$, and formulate the correlation-based prior information of the $c^{th}$ class as $[\mathbf{V}_c]_{ii} = 1 - v_{ic} \:\: i = 1,\:...\:, N_c$.
 
    Label correlation information is widely exploited to tackle the redundancy of label information in various multi-label tasks  \cite{Wang2010}, \cite{Zhu2018}. However, it can lead to a sub-optimal result, due to non-zero values in the correlation weight factor matrix for irrelevant labels \cite{Xu2018b}. Because we calculate the correlation-based prior information matrix based on each class separately, the non-zero values of unrelated label pairs can be avoided. 
    
    \subsubsection{\textit{Binary-based prior information matrix (SwMLDAb)}} Binary-based approach directly utilizes the label information as in \cite{Park2008a}. In our formulation, this approach reduces to having an equal value in $\mathbf{V}_c$ for all instances as only instances belonging to class $c$ are considered in $\mathbf{V}_c$. For wMLDA, such direct use of class labels leads to over-counting problem in the scatter matrices. In our formulation, this problem is avoided because $\mathbf{V}_c$ merely represents the prior information of non-salient instances and the final weight matrix $\mathbf{P}$ is normalized for each class.

     \subsubsection{\textit{Entropy-based prior information matrix (SwMLDAe)}} We utilize entropy metric for label information to present a prior information matrix of each class $c$, as in \cite{Xu2018b}, \cite{Chen2007}. For each instance $i$, its number of relevant labels is calculated as
    \begin{equation} \label{eq:m_i}
        m_i = \sum_{c=1}^{C}y_{ic} = {\| \mathbf{y}_i\|_{\ell_1}} < C,
    \end{equation}
    and its entropy is given as
    \begin{equation} \label{eq:m_i}
        h_i =  - \sum_{c=1}^{m_i} P_{ik} \ln(P_{ik})=  \ln(\frac{1}{m_i}),
    \end{equation}
   where $P_{ik} = 1/m_i $. Thus, the entropy is higher, when there are fewer relevant labels. The probability for an instance $i$ to be relevant to class $c$ is
    \begin{equation} \label{eq:vc_entropy}
         p^e_{ic} = e^{-h_i} = \frac{1}{m_i} = \frac{1}{\| \mathbf{y}_i\|_{\ell_1}}. 
    \end{equation}
   The entropy-based prior information of each instance $i$ to the different class(es) is defined as follows:
     \begin{equation} \label{eq:vc_entropy}
         \mathbf{v}_i = \mathbf{1} - \frac{\mathbf{y}_i}{\| \mathbf{y}_i\|_{\ell_1}}.
    \end{equation}
    Finally, the diagonal matrix $\mathbf{V}_c$ has elements $[\mathbf{V}_c]_{ii} = v_{ic}, \:\: i = 1,\:...\:, N_c$.
    
    \subsubsection{\textit{Fuzzy-based prior information matrix (SwMLDAf)}} Fuzzy $C$-means clustering algorithm (FCM) \cite{Bezdek1981} is an extension of $k$-means, where an instance can belong to multiple clusters with different degrees. The membership degree of instance $i$ in class $c$ is indicated with a weight factor $w_{ic}, \; 0<w_{ic} <1$ \cite{Dembczynski2012}. In our work, a supervised version of fuzzy $C$-means clustering algorithm (SFCM) \cite{Xu2018b}, \cite{Lin2010} is exploited to obtain the prior information matrix.
    
    As in \cite{Xu2018b}, we optimize the following:
    \begin{equation}
    \begin{split} \label{eq:vc_fuzzy}
          & \text{min}\frac{1}{2}\sum_{i=1}^{N}\sum_{c=1}^{C}w_{ic}^2 \left \| \mathbf{x}_i - \mathbf{m}_c \right \|_{\ell_2}^2, \\
          & \text{s.t.} \sum_{c=1}^{C}w_{ic}y_{ic} = 1,
    \end{split}
    \end{equation}
    where $\mathbf{m}_c$ presents the fuzzy centroid of class $c$, $w_{ic}$ denotes the membership of instance $i$ to class $c$. The constraint forces the weights of each instance $i$ to sum to one. The constrained optimization problem in Eq. (\ref{eq:vc_fuzzy}) can be solved by Lagrangian optimization with $\alpha_i \geq 0 $, where
    \begin{equation}
        \begin{split} \label{eq:vc_fuzzy_lg}
         & L = \frac{1}{2}\sum_{i=1}^{N}\sum_{c=1}^{C}w_{ic}^2 \left \| \mathbf{x}_i - \mathbf{m}_c \right \|_{\ell_2}^2 \\
         & - \sum_{i=1}^{N} \alpha_i \Big( \sum_{c=1}^{C}w_{ic}y_{ic} - 1\Big).
        \end{split}
    \end{equation}
    After getting the partial derivatives of $L$ with respect to $\mathbf{m}_c$, $w_{ik}$ and $\alpha_i$ and setting their values to zero, we get
     \begin{equation}  \label{eq:derived_m}
         \mathbf{m}_c = \frac{\sum_{i=1}^{N}w_{ic}^2 \mathbf{x}_i}{\sum_{i=1}^{N}w_{ic}^2}, 
    \end{equation}
    
    \begin{equation} \label{eq:derived_v}
         w_{ic} = \frac{\frac{y_{ic}}{\left \| \mathbf{x}_i - \mathbf{m}_c \right \|_{\ell_2}^2}}{\sum_{c=1}^{C}\frac{y_{ic}}{\left \| \mathbf{x}_i - \mathbf{m}_c \right \|_{\ell_2}^2}}.
    \end{equation}
    As the optimal value of $\mathbf{m}_c$ depends on $w_{ic}$ and vice versa, Eq. (\ref{eq:derived_m}) and Eq. (\ref{eq:derived_v}) are solved iteratively until the solution converges. Finally, we set the values of the diagonal matrix $\mathbf{V}_c$ as $[\mathbf{V}_c]_{ii} = 1 - w_{ic}, \:\: i = 1,\:...\:, N_c$.

     \subsubsection{\textit{Dependence-based prior information matrix (SwMLDAd)}} Dependence-based weights were proposed in \cite{Xu2018b}. They are based on Hilbert-Schmidt independence criterion (HSIC) \cite{Gretton2005}, which is used to describe statistical dependence between features and labels based on the estimation of Hilbert-Schmidt norms. We follow the definition of HSIC in \cite{Xu2018b} as
     \begin{equation} 
        \begin{aligned}\label{eq:hsic}
          \text{HSIC} = & \text{tr} \Big((\mathbf{Y} \circ \mathbf{W}) \mathbf{\Theta} (\mathbf{Y} \circ \mathbf{W})^\intercal\Big) \\
          = & \sum_{i,j=1}^{N_c} \theta_{ij} \Big(\mathbf{w}_i \;\text{diag} (\mathbf{y}_i \circ \mathbf{y}_j) \;\mathbf{w}_j^\intercal\Big),
        \end{aligned}
      \end{equation}
    where $\mathbf{W} =  [\mathbf{w}_1, ..., \mathbf{w}_i, ..., \mathbf{w}_{N_c}]$, $\theta_{ij} = [\mathbf{\Theta}]_{ij}$, and $\mathbf{\Theta} =  \mathbf{HX}^\intercal\mathbf{XH}$. $\mathbf{H}$ is a centered matrix, which is represented as $\mathbf{H} = \mathbf{I} - \mathbf{uu}^\intercal/N_c$, where $\mathbf{I} \in \mathbb{R}^{N_c \times N_c}$ denotes an identity matrix and $\mathbf{u} \in \mathbb{R}^{N_c}$ denotes an all-one vector. $\circ$ denotes the Hadamard, i.e., element-wise, product of two matrices or vectors. To find $\mathbf{W}$ that maximizes HSIC, we solve the following optimization problem using the iterative approach described in \cite{Xu2018b}:
    \begin{equation} 
        \begin{aligned}\label{eq:opt_hsic}
          \text{min} F(\mathbf{W}) 
          = &-\frac{1}{2} \sum_{i,j=1}^{N_c}\theta_{ij}\Big(\mathbf{w}_i\; \text{diag}(\mathbf{y}_i \circ \mathbf{y}_j)\;\mathbf{w}_j^\intercal\Big), \\
         \text{s.t.} \:\:\: \mathbf{y}_i^\intercal\mathbf{w}_i = & 1, \: \mathbf{w}_i \geq 0, \: i = 1, \;...,\; N_c .
        \end{aligned}
      \end{equation}
  This approach transforms a multi-label task to several single-label tasks \cite{Xu2018b}. It allocates 1 to only one prominent class for each instance after the final iteration. In our probabilistic formulation, the diagonal matrix $\mathbf{V}_c$ has elements $[\mathbf{V}_c]_{ii} = 1 - w_{ic}, \:\: i = 1,\:...\:, N_c$.


\section{experiments}
In our work, we tested our approach on ten multi-label databases and compared the final results with six competing methods using five evaluation metrics. We use the Matlab codes provided for \cite{Xu2018b}\;\repnote\;in the comparative experiments and exploit the relevant parts also in the implementation of our proposed method. In the following subsections, we present ten databases, implementation details, evaluation metrics, and classification results.

\subsection{Databases}
We perform our experiments on 10 publicly available multi-label databases\repnote : Yeast \cite{Nakai1992}, Scene \cite{Boutell2004}, Cal500 \cite{Turnbull2008}, Medical \cite{Pestian2007}, TMC2007-500 \cite{Srivastava2005}, Corel16k001 \cite{Barnard2003}, PlantGO\repnote , Image\repnote , HumanGO\repnote , Enron\secdnote . The contents of these databases include text, image, and acoustic clips. The numbers of classes and features of these databases are shown in Table \ref{databases}. 'Cardinality' gives the mean numbers of class labels per instance for the database.
\begin{table*}[!ht]
\centering
\caption{Characteristics of datasets used for experiments} \label{databases}
\begin{tabular}[t]{lcccccc}
\hline
Database & Contents & Train Instances& Test Instances& Classes & Attributes& Cardinality \\
\hline
Yeast &Biology &1500&917 &14 &103 &4.24 \\
PlantGO &Biology &588 &390 &12 &3091&1.08 \\
Image &Scene   &1200&800 &5  &294 &1.24 \\
Scene &Scene   &1211&1196&6  &294 &1.07 \\
Enron &Text    &1123&579 &53 &1001&3.38 \\
Cal500&Music   &300 &202  &174&68  &26.04 \\
HumanGO &Biology &1862&1244 &14 &9845&1.19 \\
Medical&Text   &645 &333  &45 &1449&1.25 \\
TMC2007-500&Text&21519&7077&22&500&2.16 \\
Corel16k001&Scene&5188&1744&153&103&4.24 \\
\hline
\end{tabular}
\end{table*}%

\subsection{Experimental setup}
 After eigendecomposition of $\mathbf{S}_{w}^{-1}\mathbf{S}_{b}$, we retained the eigenvectors corresponding to the top 0.999 informative eigenvalues. The classifier used in the experiments is a multi-label $k$-nearest neighbor classifier (ML-KNN) \cite{Zhang2007} with $k = 15$ as in \cite{Xu2018b}. ML-KNN utilizes $k$-nearest neighbor algorithm and maximum a posterior (MAP) principle to tackle the multi-label categorization task. ML-KNN first estimates prior and posterior probability of each instance $i$ for each class $c$ from a training dataset based on frequency counting \cite{Zhang2007}. Then, the predicted probabilities on a test dataset are calculated based on the prior and posterior probabilities on the training dataset using the Bayesian rule. The predicted labels are obtained by setting a threshold ($\geq 0.5$) for the predicted probabilities.
 
\subsection{Performance evaluation}
We adopt five different evaluation metrics \cite{Zhang2010}, \cite{Park2019} to evaluate the performance of our proposed algorithm: one error, normalized coverage, ranking loss, hamming loss, and macro-F1. We introduce them in the following. Here, we denote the ground truth label matrix for the $M$ test samples as $\mathbf{Y} = [\mathbf{y}_1,\:...,\:\mathbf{y}_i,\:...\:, \mathbf{y}_M]$, where the $i^{th}$ column  $\mathbf{y}_i \in \mathbb{R}^C$ represents the label vector of test sample $\mathbf{x}_i$.

The predicted label matrix is denoted as $\mathbf{\hat{Y}}$ and $\mathbf{\hat{y}}_i$ is the predicted label vector of a test sample $\mathbf{x}_i$. We use $\mathbf{\hat{p}}_i = f(\mathbf{x}_i)$ for the predicted probabilities, where $\hat{p}_{i, c} \; (0 \leq \hat{p}_{i, c} \leq 1)$ denotes the membership of instance $i$ in class $c$. $\mathcal{L}_i = \{ \text{sort}_\textit{c}(\mathbf{\hat{p}}_i) \}$ denotes an ordered list of classes ranked in the order of descending probability in $\mathbf{\hat{p}}_i$. $\mathcal{I}(\mathbf{y}_i)$ is used to denote the indices of relevant classes in $\mathbf{y}_i$ and $\neg \mathcal{I}(\mathbf{y}_i)$ denotes the indices of negative classes in $\mathbf{y}_i$.

\begin{enumerate}
    \item \textbf{\textit{One error}} shows how often the top ranked class is not among the positive ground truth labels. Lower values of this metric indicate better performance.
    
    \begin{eqnarray}\label{eq:one-error1}
          \begin{array}{r@{}l}
            one\_error_{i} =
          \end{array}
            \left\lbrace
          \begin{array}{ll}
            0,  &\text{if} \:\: \mathcal{L}_i[1] \in \mathcal{I}(\mathbf{y}_i) , \\
            1,  &\text{otherwise},
          \end{array}
          \right.
    \end{eqnarray}
    where $\mathcal{L}_i[1]$ denotes the first class in the sorted list $\mathcal{L}_i$.
    \begin{equation} \label{eq:one_error}
        one\_error = \frac{\sum_{i=1}^{M} one\_error_{i}}{M}. 
    \end{equation}
  
    \item \textbf{\textit{Normalized coverage}} demonstrates how far on average in the predicted label ranking $\mathcal{L}_i$ one needs to go to cover all the ground-truth labels of an instance. A smaller coverage value indicates better performance.
  
    \begin{gather}\label{eq:coverage}
      coverage = \frac{\sum_{i=1}^{M} \text{max}_j \{j| \mathcal{I}(\mathbf{y}_i)  \in_j \mathcal{L}_i \} - 1}{M * (C-1)},
    \end{gather}
  where $\{j| \mathcal{I}(\mathbf{y}_i) \in_j \mathcal{L}_i \}$ gives the positions of relevant classes $\mathcal{I}(\mathbf{y}_i)$  in the ordered list $\mathcal{L}$.
    \item \textbf{\textit{Ranking loss}} evaluates for each item $i$ relevant vs. irrelevant class pair and gives the fraction of pairs, where the irrelevant class if ranked above the relevant one. Smaller values of this metric indicate a better performance. Here, we use $m$ to denote the number of relevant classes in $\mathbf{y}_i$ and $n = C-m$:
     
    \begin{gather}\label{eq:ranking_loss}
      ranking\_loss_{i} =  \frac{| \hat{p}_{i, \mathcal{I}(\mathbf{y}_i)} \leq \; \hat{p}_{i, \neg \mathcal{I}(\mathbf{y}_i)} |}{m*n}, \\
      ranking\_loss = \frac{\sum_{i=1}^{M} ranking\_loss_{i}}{M},
    \end{gather}
    where $| \hat{p}_{i, \mathcal{I}(\mathbf{y}_i)} \leq \; \hat{p}_{i, \neg \mathcal{I}(\mathbf{y}_i)} |$ is used to denote the count of wrong rankings for item $i$. 
    \item \textbf{\textit{Hamming loss}} shows the rate of misclassified predicted values using XOR comparison between predicted labels and ground truth labels. Smaller values of this metric indicate a better performance:
    \begin{equation} \label{eq:one_error}
        hamming\_loss = \frac{1}{M} \sum_{i=1}^{M} \frac{\|\mathbf{y}_i \:\: \oplus \:\: \mathbf{\hat{y}}_i  \|_{\ell_1}}{C}.
    \end{equation}
    
    
    
    \item \textbf{\textit{Macro-F1}} shows the average F1 value on each class, which reveals the authenticity and reliability of predicted true labels. Higher values of this metric indicate a better performance.
    
    \begin{gather} \label{eq:MACRO-F1}
        macro\-F1 = \frac{2}{C} \sum_{c=1}^{C} \frac{precision_c * recall_c}{precision_c + recall_c},
    \end{gather}
    where $precision_c$ and $recall_c$ are precision and recall for class $c$.
\end{enumerate}

\begin{table*}[!ht]
\caption{One error ($\downarrow$)}
\label{one_error}
\resizebox{\textwidth}{!}{
\begin{tabular}{c c c c c c c | c c c c c c}
\hline
&\multicolumn{6}{c|}{Reference methods} & \multicolumn{6}{c}{Variants of the proposed saliency-based methods} \\
Dataset & DMLDA & wMLDA\textsubscript{c} & wMLDA\textsubscript{b} & wMLDA\textsubscript{e} & wMLDA\textsubscript{f} & wMLDA\textsubscript{d} &SwMLDA\textsubscript{m} & SwMLDA\textsubscript{c} & SwMLDA\textsubscript{b} & SwMLDA\textsubscript{e} & SwMLDA\textsubscript{f} & SwMLDA\textsubscript{d} \\
\hline
Yeast       &0.2399&0.2486&0.2410&0.2475&0.2530&0.2497&0.2474&0.2530&0.2530&0.2421&0.2432&$\boldsymbol{0.2257}$\\
Plant       &0.7564&0.7359&0.6069&0.7436&0.7359&0.7410&$\boldsymbol{0.6436}$&0.6692&0.6615&0.6667&0.6564&0.6590\\
Image       &0.4975&0.3413&0.3613&0.3400&0.3463&0.3463&0.3325&$\boldsymbol{0.3063}$&0.3150&0.3163&0.3263&0.3213\\
Scene       &0.4983&0.3286&0.3202&0.3269&0.3286&0.3202&0.2542&$\boldsymbol{0.2400}$&0.2408&$\boldsymbol{0.2400}$&0.2425&0.2416\\
Enron       &0.7636&0.8061&0.7242&0.7000&0.6909&0.5924&0.5348&0.5833&0.5833&0.5455&0.5576&$\boldsymbol{0.5000}$\\
Cal500      &$\boldsymbol{0.1040}$&$\boldsymbol{0.1040}$&0.1089&$\boldsymbol{0.1040}$&$\boldsymbol{0.1040}$&0.1386&$\boldsymbol{0.1040}$&0.1139&0.1089&0.1139&0.1139&$\boldsymbol{0.1040}$\\
Human       &0.6849&0.6174&0.6069&0.6174&0.6094&0.5997&$\boldsymbol{0.5804}$&0.6109&0.6109&0.6045&0.6029&0.5916\\
Medical     &0.3964&0.2613&0.2252&0.2342&0.2222&0.2312&0.2162&0.2012&0.2042&0.1922&0.1922&$\boldsymbol{0.1892}$\\
TMC2007     &0.2021&0.1498&0.1492&0.1495&$\boldsymbol{0.1471}$&0.1499&0.1584&0.1561&0.1557&0.1553&0.1537&0.1501\\
Corel16k001 &0.7414&0.7259&0.7242&0.7288&0.7208&0.7104&$\boldsymbol{0.7047}$&0.7299&0.7185&0.7150&0.7150&0.7225\\
\hline
\end{tabular}}
\end{table*}

\begin{table*}[!ht]
\caption{Normalized coverage ($\downarrow$)}\label{Coverage}
\resizebox{\textwidth}{!}{
\begin{tabular}{c c c c c c c | c c c c c c}
\hline
&\multicolumn{6}{c|}{Reference methods} & \multicolumn{6}{c}{Variants of the proposed saliency-based methods} \\
Dataset & DMLDA & wMLDA\textsubscript{c} & wMLDA\textsubscript{b} & wMLDA\textsubscript{e} & wMLDA\textsubscript{f} & wMLDA\textsubscript{d} &SwMLDA\textsubscript{m} & SwMLDA\textsubscript{c} & SwMLDA\textsubscript{b} & SwMLDA\textsubscript{e} & SwMLDA\textsubscript{f} & SwMLDA\textsubscript{d} \\
\hline
Yeast       &0.5187  &0.5119  &0.5072  &0.5012  &0.5003  &0.5097  &$\boldsymbol{0.4951}$&0.4991  &0.4987 &0.4962  &0.4975  &0.4964\\
Plant       &0.2646  &0.2846  &0.2355  &0.2900  &0.2984  &0.2797  &$\boldsymbol{0.2277}$&0.2303  &0.2387 &0.2282  &0.2359  &0.2408\\
Image       &0.3528  &0.2619  &0.2641  &0.2656  &0.2659  &0.2656  &0.2416  &0.2313  &$\boldsymbol{0.2222}$&0.2284  &0.2300  &0.2269\\
Scene       &0.2547  &0.1584  &0.1574  &0.1547  &0.1567  &0.1515  &0.1112  &0.1110  &0.1110 &$\boldsymbol{0.1103}$&0.1109  &0.1139\\
Enron       &0.3457  &0.3862  &0.3650  &0.3479  &0.3545  &0.3361  &$\boldsymbol{0.3024}$&0.3043  &0.3058 &0.3095  &0.3073  &0.3032\\
Cal500      &$\boldsymbol{0.7433}$&0.7533  &0.7477  &0.7511  &0.7517  &0.7486  &0.7472  &0.7462  &0.7467 &0.7468  &0.7469  &0.7469\\
Human       &0.2127  &0.1969  &0.1945  &0.1964  &0.1971  &0.1945  &$\boldsymbol{0.1819}$&0.1855  &0.1855 &0.1834  &0.1845  &0.1835\\
Medical     &0.0819  &0.0779  &0.0819  &0.0704  &0.0716  &0.0762  &$\boldsymbol{0.0607}$&0.0678  &0.0659 &0.0678  &0.0665  &0.0634\\
TMC2007     &0.1148  &0.0983  &0.0974  &0.0979  &0.0973  &0.1024  &0.0994  &0.0970  &$\boldsymbol{0.0966}$&0.0972  &0.0973  &0.0975\\
Corel16k001 &0.3956  &0.3771  &0.3698  &0.3740  &0.3731  &0.3779  &$\boldsymbol{0.3574}$&0.3698  &0.3677 &0.3687  &0.3698  &0.3639\\
\hline
\end{tabular}}
\end{table*}

\begin{table*}[!ht]
\caption{Ranking loss ($\downarrow$) }\label{ranking_loss}
\resizebox{\textwidth}{!}{
\begin{tabular}{c c c c c c c | c c c c c c}
\hline
&\multicolumn{6}{c|}{Reference methods} & \multicolumn{6}{c}{Variants of the proposed saliency-based methods} \\
Dataset & DMLDA & wMLDA\textsubscript{c} & wMLDA\textsubscript{b} & wMLDA\textsubscript{e} & wMLDA\textsubscript{f} & wMLDA\textsubscript{d} &SwMLDA\textsubscript{m} & SwMLDA\textsubscript{c} & SwMLDA\textsubscript{b} & SwMLDA\textsubscript{e} & SwMLDA\textsubscript{f} & SwMLDA\textsubscript{d} \\
\hline
Yeast       &0.1900  &0.1827  &0.1823  &0.1799  &0.1808  &0.1813  &0.1744  &0.1786 &0.1777  & 0.1761 & 0.1748 &$\boldsymbol{0.1724}$\\
Plant       &0.2577  &0.2763  &$\boldsymbol{0.1737}$  &0.2817  &0.2878  &0.2713  &0.2196  &0.2254 &0.2300  & 0.2199 & 0.2274 &0.2315\\
Image       &0.2878  &0.1948  &0.1978  &0.1986  &0.2000  &0.1992  &0.1771  &0.1599 &$\boldsymbol{0.1588}$  & 0.1653 & 0.1667 &0.1652\\
Scene       &0.2321  &0.1380  &0.1367  &0.1338  &0.1360  &0.1318  &0.0909  &$\boldsymbol{0.0890}$ &0.0892  & 0.0896 & 0.0900 &0.0929\\
Enron       &0.1739  &0.2012  &0.1742  &0.1639  &0.1682  &0.1537  &0.1306  &0.1330 &0.1330  & 0.1336 & 0.1329 &$\boldsymbol{0.1279}$\\
Cal500      &0.1882  &0.1900  &0.1882  &0.1865  &0.1863  &$\boldsymbol{0.1842}$  &0.1854  &0.1860 &0.1854  & 0.1854 & 0.1855 &0.1865\\
Human       &0.1907  &0.1712  &0.1702  &0.1712  &0.1721  &0.1702  &$\boldsymbol{0.1584}$  &0.1602 &0.1612  & 0.1604 & 0.1609 &0.1603\\
Medical     &0.0682  &0.0571  &0.0648  &0.0527  &0.0498  &0.0570  &$\boldsymbol{0.0445}$  &0.0462 &0.0480  & 0.0489 & 0.0482 &0.0461\\
TMC2007     &0.0375  &0.0269  &0.0266  &0.0268  &0.0264  &0.0289  &0.0279  &$\boldsymbol{0.0261}$ &$\boldsymbol{0.0261}$  & 0.0264 & 0.0264 &0.0263\\
Corel16k001 &0.1962  &0.1894  &0.1863  &0.1872  &0.1864  &0.1890  &$\boldsymbol{0.1796}$  &0.1866 &0.1857  & 0.1863 & 0.1868 &0.1825\\
\hline
\end{tabular}}
\end{table*}

\begin{table*}[!ht]
\caption{Hamming loss ($\downarrow$)}\label{hamming loss}
\resizebox{\textwidth}{!}{
\begin{tabular}{c c c c c c c | c c c c c c}
\hline
&\multicolumn{6}{c|}{Reference methods} & \multicolumn{6}{c}{Variants of the proposed saliency-based methods} \\
Dataset & DMLDA & wMLDA\textsubscript{c} & wMLDA\textsubscript{b} & wMLDA\textsubscript{e} & wMLDA\textsubscript{f} & wMLDA\textsubscript{d} &SwMLDA\textsubscript{m} & SwMLDA\textsubscript{c} & SwMLDA\textsubscript{b} & SwMLDA\textsubscript{e} & SwMLDA\textsubscript{f} & SwMLDA\textsubscript{d} \\
\hline
Yeast       &0.2077&0.2046&0.2028&0.2035&0.2049&0.2091&$\boldsymbol{0.2003}$&0.2038&0.2059&0.2047&0.2046&0.2049\\
Plant       &$\boldsymbol{0.0921}$&0.1171&0.0924&0.1184&0.1201&0.1081&0.0947&0.1017&0.1010&0.1021&0.1068&0.0987\\
Image       &0.2310&0.1893&0.1898&0.1860&0.1883&0.1828&0.1738&0.1703&0.1723&0.1698&0.1713&$\boldsymbol{0.1683}$\\
Scene       &0.1683&0.1182&0.1185&0.1198&0.1256&0.1172&0.0975&0.0917&0.0917&0.0949&$\boldsymbol{0.0914}$&0.0943\\
Enron       &0.0669&0.0721&0.0668&0.0645&0.0664&$\boldsymbol{0.0548}$&0.0565&0.0585&0.0585&0.0563&0.0565&0.0549\\
Cal500      &0.1392&0.1394&0.1393&0.1386&0.1383&0.1391&0.1390&0.1398&$\boldsymbol{0.1381}$&0.1388&0.1386&0.1383\\
Human       &$\boldsymbol{0.0843}$&0.0943&0.0908&0.0924&0.0923&0.0908&0.0845&0.0891&0.0887&0.0868&0.0880&0.0874\\
Medical     &0.0225&0.0172&0.0225&0.0167&0.0161&0.0165&0.0159&0.0153&0.0149&0.0153&0.0155&$\boldsymbol{0.0146}$\\
TMC2007     &0.0608&0.0539&0.0529&0.0535&0.0531&0.0571&0.0544&0.0537&0.0531&0.0535&0.0535&$\boldsymbol{0.0520}$\\
Corel16k001 &0.0200&0.0200&$\boldsymbol{0.0199}$&0.0200&$\boldsymbol{0.0199}$&0.0200&0.0201&0.0200&0.0200&0.0200&0.0200&0.0200\\
\hline
\end{tabular}}
\end{table*}

\begin{table*}[!ht]
\caption{Macro-F1 ($\uparrow$)}\label{Macro-F1}
\resizebox{\textwidth}{!}{
\begin{tabular}{c c c c c c c | c c c c c c}
\hline
&\multicolumn{6}{c|}{Reference methods} & \multicolumn{6}{c}{Variants of the proposed saliency-based methods} \\
Dataset & DMLDA & wMLDA\textsubscript{c} & wMLDA\textsubscript{b} & wMLDA\textsubscript{e} & wMLDA\textsubscript{f} & wMLDA\textsubscript{d} &SwMLDA\textsubscript{m} & SwMLDA\textsubscript{c} & SwMLDA\textsubscript{b} & SwMLDA\textsubscript{e} & SwMLDA\textsubscript{f} & SwMLDA\textsubscript{d} \\
\hline
Yeast       &0.3174&0.3516&0.3596&0.3532&$\boldsymbol{0.3700}$&0.2988&0.3519&0.3342&0.3486&0.3483&0.3475&0.3647\\
Plant       &0.0185&0.1259&0.1574&0.1216&0.1543&0.1331&0.1461&0.1488&$\boldsymbol{0.1619}$&0.1503&0.1583&0.1393\\
Image       &0.3002&0.5908&0.5738&0.5852&0.5875&0.5774&0.5610&0.5854&0.5956&$\boldsymbol{0.5998}$&0.5864&0.5686\\
Scene       &0.3456&0.6488&0.6523&0.6412&0.6406&0.6489&0.7106&$\boldsymbol{0.7322}$&0.7306&0.7269&0.7304&0.7294\\
Enron       &0.0198&0.0372&0.0483&0.0600&0.0557&0.0331&0.0637&0.0567&0.0524&0.0595&0.0595&$\boldsymbol{0.0714}$\\
Cal500      &0.0526&$\boldsymbol{0.0553}$&0.0465&0.0504&0.0501&0.0525&0.0490&0.0520&0.0542&0.0527&0.0522&0.0511\\
Human       &0.0016&$\boldsymbol{0.1487}$&0.1460&0.1493&0.1455&0.1460&0.1380&0.1303&0.1371&0.1300&0.1429&0.1431\\
Medical     &0.1302&0.1911&0.1302&0.1959&0.1898&0.1916&0.1921&0.2253&0.2043&0.2210&0.2222&$\boldsymbol{0.2247}$\\
TMC2007     &0.4748&0.5917&0.5994&0.5921&0.5928&0.5394&0.6120&0.6125&$\boldsymbol{0.6202}$&0.6022&0.6074&0.6147\\
Corel16k001 &0.0184&0.0353&0.0373&0.0305&0.0304&0.0315&$\boldsymbol{0.0489}$&0.0361&0.0379&0.0386&0.0366&0.0447\\
\hline
\end{tabular}}
\end{table*}

\subsection{Classification results}
Tables (\ref{one_error})-(\ref{Macro-F1}) show the experimental results of our approach and competing methods with one error, normalized coverage, ranking loss, hamming loss, and macro-F1 metrics. One error, normalized coverage, and ranking loss directly utilize the probabilities from the ML-KNN algorithm in various ways. We can conclude that all versions of our proposed methods achieved significant improvements in most databases comparing to the reference methods with the first three metrics that use probabilities. Our method achieved the best result in eight cases out of ten in Tables (\ref{one_error}) and (\ref{ranking_loss}), and nine cases out of then in Table (\ref{Coverage}). 

The remaining two metrics utilize the predicted labels obtained by a threshold value $0.5$ and the probabilities in different ways. We currently did not adapt a cross validation strategy to select an optimal threshold value in our experiments, which may lead to suboptimal results. With the last two metrics, the reference methods worked a bit better than with the former three metrics, but even in the worst case with Hamming loss, our method achieved the best results in six cases out of ten. 

According to the results with all metrics, our mis-classification-based prior information variant $SwMLDA_m$ is the most efficient and precise one and totally achieved 15 best results among all the test cases (highlighted values in the tables). $SwMLDA_d$ achieved 11 best results among all the test cases with different metrics. Moreover, each variant of our algorithm achieved better performance compared to the corresponding reference methods in most cases. For instance, $SwMLDA_f$ achieved better results on at least eight cases out of ten for any metric than $wMLDA_f$ did with the one-error metric, $SwMLDA_d$ enhanced the performance on eight cases with hamming loss and seven cases with marco-F1 compared to $wMLDA_d$. This shows that the proposed approach of using the prior information for class saliency estimation generally outperforms using it directly for weighting the items as in \cite{Xu2018b}.

\section{Conclusion}
In this paper, we proposed a novel multi-label classification method to tackle the data imbalance and information redundancy problems in encountered multi-label classification tasks. Our method is an extension MLDA, where the weights are generated with a probabilistic approach to evaluate the saliency of each instance for different classes. The probabilistic approach uses an affinity matrix to ensure similar results for similar instances and a prior information matrix to integrate prior information on prominence of each instance for each class. Our solution can alleviate the data imbalance problem, which is commonly encountered in multi-label databases, as the weight factor vectors are calculated separately for each class. Our method can also alleviates the common over-counting problem. We proposed variants of our methods using different prior information matrices based on both labels and features. 

We used five metrics to evaluate the performance of our method with competing method on ten multi-label datasets. The experimental results show that our method enhanced the classification performance compared to the competing algorithms.

Our algorithm is still based on the linear subspace learning technique. In the future, we will make a non-linear extension using the kernel trick. We will also explore the prominence of each feature channel from all instances to calculate the weight factor vector.


%





\ifCLASSOPTIONcaptionsoff
  \newpage
\fi





\bibliographystyle{IEEEtran}
\bibliography{references}
%

\begin{IEEEbiographynophoto}
{Lei Xu} received the B.S.E.E degree from East China Normal University, ShangHai, China, in 2006, the M.S.E.E. degree from the University of Tampere, Finland, in 2017. From 2006 to 2013, she was an Engineer in ShangHai, where she was involved with on-train communication systems design. Her current research interests include Artificial Intelligence, or/and Machine Learning. She is currently a PhD candidate at Tampere University.\end{IEEEbiographynophoto}

\begin{IEEEbiographynophoto}
{Jenni Raitoharju} received her Ph.D. degree at Tampere University of Technology, Finland in 2017. Since then, she has worked as a Postdoctoral Research Fellow at the Faculty of Information Technology and Communication Sciences, Tampere University, Finland. In 2019, she started working as a Senior Research Scientist at the Finnish Environment Institute, Jyväskylä, Finland after receiving Academy of Finland Postdoctoral Researcher funding for 2019-2022. She has co-authored 12 journal papers and 24 papers in international conferences. She is the chair of Young Academy Finland 2019-2020. Her research interests include machine learning and pattern recognition methods along with applications in biomonitoring and autonomous systems.
\end{IEEEbiographynophoto}

\begin{IEEEbiographynophoto}
{Alexandros Iosifidis} (SM'16) is an Associate Professor at Aarhus University, Denmark. He has (co-)authored 65 articles in international journals and 85 papers in international conferences proposing novel Machine Learning techniques and their application in a variety of problems. He served as an Officer of the Finnish IEEE SP/CAS Chapter (2016-2018), he is a member of the EURASIP Technical Area Committee on Visual Information Processing, and serves as Associate Editor for Neurocomputing, Signal Processing: Image Communications, and BMC Bioinformatics journals. His research interests include topics of neural networks and statistical machine learning finding applications in computer vision, financial engineering and graph mining.
\end{IEEEbiographynophoto}

\begin{IEEEbiographynophoto}{Moncef Gabbouj}
received his BS degree in 1985 from Oklahoma State University, and his MS and PhD degrees from Purdue University, in 1986 and 1989, respectively, all in electrical engineering. Dr. Gabbouj is a Professor of Signal Processing at the Department of Computing Sciences, Tampere University, Tampere, Finland. He was Academy of Finland Professor during 2011-2015. His research interests include Big Data analytics, multimedia content-based analysis, indexing and retrieval, artificial intelligence, machine learning, pattern recognition, nonlinear signal and image processing and analysis, voice conversion, and video processing and coding. Dr. Gabbouj is a Fellow of the IEEE and member of the Academia Europaea and the Finnish Academy of Science and Letters. He is the past Chairman of the IEEE CAS TC on DSP and committee member of the IEEE Fourier Award for Signal Processing. He served as associate editor and guest editor of many IEEE, and international journals and Distinguished Lecturer for the IEEE CASS. Dr. Gabbouj is the Finland Site Director of the NSF IUCRC funded Center for Visual and Decision Informatics (CVDI) and leads the Artificial Intelligence Research Task Force of the Ministry of Economic Affairs and Employment funded Research Alliance on Autonomous Systems (RAAS).\end{IEEEbiographynophoto}




\vfill
\end{document}